\documentclass{article} 
\usepackage{iclr2022_conference,times}

\usepackage{amsmath,amsfonts,bm}

\def\eqref#1{equation~\ref{#1}}

\def\1{\bm{1}}

\DeclareMathAlphabet{\mathsfit}{\encodingdefault}{\sfdefault}{m}{sl}
\SetMathAlphabet{\mathsfit}{bold}{\encodingdefault}{\sfdefault}{bx}{n}

\usepackage{hyperref}
\usepackage{url}

\usepackage[utf8]{inputenc} 
\usepackage[T1]{fontenc}    
\usepackage{hyperref}       
\usepackage{url}            
\usepackage{booktabs}       
\usepackage{amsfonts}       
\usepackage{nicefrac}       
\usepackage{microtype}      
\usepackage{xcolor}         
\usepackage{cprotect} 

\usepackage{amsmath}
\usepackage{cleveref}
\usepackage{import}

\usepackage{hyperref}

\usepackage{color}
\definecolor{keywordcolor}{rgb}{0.7, 0.1, 0.1}   
\definecolor{tacticcolor}{rgb}{0.1, 0.2, 0.6}    
\definecolor{commentcolor}{rgb}{0.4, 0.4, 0.4}   
\definecolor{symbolcolor}{rgb}{0.0, 0.1, 0.6}    
\definecolor{sortcolor}{rgb}{0.1, 0.5, 0.1}      
\definecolor{attributecolor}{rgb}{0.7, 0.1, 0.1} 

\usepackage{listings}

\usepackage{stmaryrd}

\usepackage{upquote} 
\usepackage{amssymb} 
\usepackage{mathtools} 

\usepackage{color}
\usepackage{listings}

\newcommand\ie{\emph{i.e.}\ }
\newcommand\eg{\emph{e.g.}\ }

\usepackage{array,etoolbox}
\preto\tabular{\setcounter{magicrownumbers}{0}}
\newcounter{magicrownumbers}

\title{Proof Artifact Co-training for Theorem Proving with
	Language Models}

\author{%
	Jesse Michael Han \\
	University of Pittsburgh \\
	OpenAI \\
	\And
	Jason Rute \\
	IBM Research\thanks{Work performed while Jason Rute was at CIBO Technologies.} \\
	\And
	Yuhuai Wu \\
	Google Research \\
	Stanford University\thanks{Work performed while Yuhuai Wu was at University of Toronto.} \\
	\AND
	Edward W. Ayers \\
    Carnegie Mellon University\thanks{Work performed while Edward W. Ayers was at University of Cambridge.}\\
	\And
	Stanislas Polu \\
	OpenAI \\
}

\iclrfinalcopy 
\begin{document}

\maketitle

\begin{abstract}
Labeled data for imitation learning of theorem proving in large
libraries of formalized mathematics is scarce, as such libraries
require years of concentrated effort by human specialists to be
built. This is particularly challenging when applying large Transformer
language models to tactic prediction, because the scaling of
performance with respect to model size is quickly disrupted in the
data-scarce, easily-overfitted regime.  We propose PACT ({\bf P}roof
{\bf A}rtifact {\bf C}o-{\bf T}raining),
a general methodology for extracting abundant self-supervised
data from kernel-level proof terms for joint training alongside the usual
tactic prediction objective.
We apply this methodology to Lean, a proof assistant host to some of the most sophisticated
formalized mathematics to date. We instrument Lean with a neural
theorem prover driven by a Transformer language model and show that
PACT improves theorem proving success rate on a
held-out suite of test theorems from 32\% to 48\%.

\end{abstract}

\section{Introduction}\label{sec:intro}


Deep learning-driven automated theorem proving in large
libraries of formalized mathematics (henceforth ``neural theorem
proving'') has been the focus of increased attention in recent
years. Labeled data for imitation
learning of theorem proving is scarce---formalization is notoriously
labor-intensive, with an estimated cost of 2.5 man-years per megabyte
of formalized
mathematics~\citep{wiedijk}, and complex projects require years of
labor from human specialists. Within a fixed corpus of (possibly unproven)
theorem statements, it is possible to augment a seed dataset of human
proofs with new successful trajectories using reinforcement learning or
expert iteration. However, for some large models 
this can be quite computationally intensive,
and without a way to expand the curriculum of theorems, the agent will
inevitably saturate and suffer from data starvation.

Data scarcity is a particularly thorny obstruction for applying large
language models (LLMs) to neural theorem proving. LLMs have
achieved spectacular success in data-rich regimes such as plain
text~\citep{DBLP:conf/nips/BrownMRSKDNSSAA20},
images~\citep{DBLP:journals/corr/abs-2010-11929}, and joint text-image
modeling~\citep{radford2learning}, and the performance
of decoder-only Transformers has been empirically shown to obey
scaling power laws in model and data
size~\citep{DBLP:journals/corr/abs-2010-14701}. However, existing datasets of human
proof steps for neural theorem proving are extremely small and exist at scales
at which overfitting occurs extremely rapidly, disrupting the scaling
of performance with respect to model
size~\citep{DBLP:journals/corr/abs-2001-08361}.


We make two contributions towards addressing the problem of data
scarcity in the context of formal mathematics. First, we introduce PACT ({\bf
P}roof {\bf A}rtifact {\bf C}o-{\bf T}raining), a general methodology for
extracting self-supervised auxiliary tasks for jointly training a language model
alongside a tactic prediction objective for interactive theorem
proving. Second, we present {\sc LeanStep}, a collection of datasets
and a machine learning environment for the Lean 3 theorem prover with
support for PACT, supervised learning of tactic prediction, theorem
proving evaluation, and reinforcement learning.


We train large language models on these data and demonstrate that
PACT significantly improves
theorem proving success rate on a held-out suite of test theorems,
from 32\% to 48\%. We then embark on a careful study of the effects of
pre-training vs. co-training and show that PACT combined with {\em WebMath}
pre-training~\citep{DBLP:journals/corr/abs-2009-03393}
achieves the best validation loss and theorem proving success rate.
Finally, on an out-of-distribution collection of thousands of theorems
(some involving novel definitions) added to Lean's mathematical library after
we extracted our train/test data, we achieve a theorem proving success
rate of \(37\%\), suggesting strong generalization and usefulness at
the frontier of formalized mathematics.








\section{Background and related work}\label{sec:background}

\subparagraph*{Lean} Lean is an interactive theorem prover and
functional programming language~\citep{DBLP:conf/cade/MouraKADR15}. It
has an extremely active community and is host to some of the most
sophisticated formalized mathematics in the world, including scheme
theory~\citep{buzzard2019schemes}, forcing~\citep{DBLP:conf/cpp/HanD20},
perfectoid spaces~\citep{DBLP:conf/cpp/BuzzardCM20}, and condensed
mathematics~\citep{lean-liquid}. Lean's foundational logic is a
dependent type theory called the
calculus of inductive constructions~\citep{DBLP:conf/mfps/PfenningP89}.
This design means that terms, types and proofs are all represented
with a single datatype called an
\emph{expression}.
A \emph{proof term} is a Lean expression whose type is
a proposition, \ie a theorem. This proof term serves as a checkable artifact for
verifying the proposition. Lean uses a small,
trusted kernel to verify proof terms.
The primary repository of formalized
mathematics in Lean is
\verb`mathlib`
~\citep{DBLP:conf/cpp/X20}.  At the time of writing, 140 contributors
have added almost 500,000 lines of code; \verb`mathlib` contains
over 46,000 formalized lemmas backed by over 21,000
definitions, 
covering topics such as
algebraic geometry, computability, measure theory, and
category theory. The range of topics and the
monolithic, unified organization of
\verb`mathlib` make it an excellent
foundation for 
a neural theorem proving dataset.

\subparagraph{Machine learning in interactive theorem proving}

In a tactic-based interactive theorem prover (ITP) such as Lean,
a proof is a list of tactics, \ie small proof-term-generating programs.  Tactics
can be simple one-word commands, \eg \lstinline|refl|, or be composed
of many nested parts, \eg
\begin{lstlisting}
    simpa [le_antisymm_iff, norm_nonneg] using @norm_eq_zero α _ g
\end{lstlisting}
Here the brackets enclose a list of simplifier rules (which often are just lemmas from the library),
and \lstinline|@norm_eq_zero α _ g| is a proof
term applying the lemma \lstinline|norm_eq_zero| to the local
variables \lstinline|α| and \lstinline|g|.

Other ML and neural theorem provers for tactic-based ITPs take one of two
approaches to tactic generation.
TacticToe~\citep{DBLP:journals/corr/abs-1804-00596} for HOL4
and Tactician~\citep{DBLP:conf/mkm/BlaauwbroekUG20} for Coq
use k-NN to select similar tactics in the training set and apply modifications
to the result, \eg swapping the tactic variables with those found in the local context.
HOList/DeepHOL~\citep{DBLP:conf/icml/BansalLRSW19,
bansal2019learning, DBLP:conf/aaai/PaliwalLRBS20}
for HOL Light;
TacticZero~\citep{DBLP:journals/corr/abs-2102-09756}
for HOL4; and
CoqGym/ASTactic~\citep{DBLP:conf/icml/YangD19} and
ProverBot9001~\citep{DBLP:conf/pldi/Sanchez-SternAS20} for Coq
hard-code the DSL for every tactic command.  The model chooses
a tactic command, and then fills in the tactic arguments using specialized
argument selectors (such as a lemma selector, a local hypothesis selector,
and/or a variable selector).
None of these selectors currently synthesize arbitrary
terms.  This prevents the tactic synthesis from constructing
tactics with proof terms, such as \lstinline|@norm_eq_zero α _ g|,
or directly proving
an existential, \eg \lstinline|∃ (x : ℝ), x + 3 = 0|, by supplying
the witnessing term \lstinline|-3|.

Directly applying generative language modeling to tactic
generation allows this setup to be considerably simplified.
Our tactic generator is able to synthesize tactics of any form found in
\texttt{mathlib} including, for example, the \lstinline|simpa| example above
as a one line proof to a test theorem,
even though the string \lstinline|@norm_eq_zero| does not occur in our dataset.
(See more examples in \Cref{sec:appendix_examples}.)  We leave as future work the
possibility of re-integrating specialized components,
\eg lemma selection, found in other works
(possibly as, say, a source of additional prompts for the language model).

Language models have also been explored in the first-order ITP
Mizar for conjecturing and proof synthesis~\citep{DBLP:conf/mkm/UrbanJ20}.
While their work shows the promise of such approaches,
is not intended as a complete end-to-end theorem prover.
For Metamath, which does not use tactics,
language modeling approaches have been quite successful.
Holophrasm~\citep{DBLP:journals/corr/Whalen16},
MetaGen~\citep{DBLP:conf/nips/WangD20},
and GPT-f~\citep{DBLP:journals/corr/abs-2009-03393} all use RNNs or Transformers
to generate proof steps.  Indeed, our paper builds on the work of Metamath
GPT-f~\citep{DBLP:journals/corr/abs-2009-03393} (MM GPT-f).
Whereas MM GPT-f trained primarily on the Metamath proof step objective (\ie
guessing the next lemma to be applied to a goal, which is similar to our
\texttt{NEXTLEMMA} task in~\Cref{subsec:pact}), we co-train
on a diverse suite of self-supervised tasks extracted from Lean proof
terms and demonstrate significant improvements in theorem proving
performance when doing so.  This is our main result.

%
%

\subparagraph{Reasoning with Transformers}
Besides theorem proving, a number of recent papers have shown that
language models, especially Transformers, are capable of
something like mathematical and logical reasoning in
integration~\citep{DBLP:conf/iclr/LampleC20},
differential equations~\citep{DBLP:journals/corr/abs-2006-06462},
Boolean satisfiability~\citep{DBLP:journals/corr/abs-2003-04218},
and inferring missing proof steps~\citep{li2021isarstep}.


A closely-related vein of work has shown that pre-training
Transformers on data engineered to reflect inductive
biases conducive to mathematical reasoning is beneficial for
downstream mathematical reasoning tasks
\citep{DBLP:journals/corr/abs-2006-04757, DBLP:journals/corr/abs-2101-06223}.
Our work both builds on and departs from these ideas in several
ways. Unlike skip-tree training
\citep{DBLP:journals/corr/abs-2006-04757}, which focuses solely on
predicting masked subterms of theorem {\em statements}, PACT derives its
self-supervised training data from far more complex {\em
  proofs}. Unlike LIME \citep{DBLP:journals/corr/abs-2101-06223},
which uses purely synthetic data and is
presented as a pre-training methodology, our self-supervised tasks are
extracted from non-synthetic human proofs. Moreover, we show that not only are Transformers
capable of performing well on auxiliary tasks gathered from
low-level proof artifact data,
but that we can directly leverage this low-level data by jointly
training a language model to greatly improve its performance at
high-level theorem proving.

\subparagraph{Machine learning with proof artifacts}
The idea of mining low-level proof artifacts was previously explored by
Kaliszyk and Urban in the context of automated lemma
extraction~\citep{DBLP:journals/jsc/KaliszykU15, DBLP:conf/frocos/KaliszykUV15}. %
It has also been previously observed that training on fully elaborated
Coq terms~\citep{DBLP:conf/cade/NiePLG20} helps with a downstream
theorem naming task. However,
similar to previous work on skip-tree training, their dataset focuses
solely on theorem statements, \ie types, does not cover the far
more complex proof terms, and does not evaluate the effect of such
training on theorem proving evaluations.

While there exist environments and datasets for other formal mathematics
libraries~\citep{DBLP:conf/iclr/KaliszykCS17,li2021isarstep,
DBLP:journals/corr/abs-1806-00608, DBLP:journals/jar/KaliszykU15a},
{\sc LeanStep} is the first and only tactic proof dataset for the Lean theorem prover.
This makes available a large set of formal mathematical data to researchers
covering a diverse and deep spectrum of pure mathematics.
Moreover,  {\sc LeanStep} is unique in that it contains both
high-level human-written tactics
as well as kernel-level proof terms, which enables the extraction of
self-supervised tasks for PACT (\Cref{subsec:pact}).


\section{The {\sc LeanStep} datasets and machine learning environment}\label{sec:embed}

\subsection{Human tactic proof steps}\label{subsec:tactic}



Tactics in Lean are
metaprograms~\citep{DBLP:journals/pacmpl/EbnerURAM17}, which can
construct Lean expressions, such as proof terms. 
A {\em tactic state} which tracks the list of open goals and other
metadata (like the partial proof term constructed
so far) is threaded through each tactic invocation.
Lean has special support for treating tactics as an extensible
domain-specific language (DSL); this DSL is how Lean is typically used
as an interactive theorem prover. The DSL amounts to a linear chain of
comma-separated invocations. 
The Lean {\em proof step} task is to predict the next tactic given
this goal state.  
We refer the reader to \Cref{sec:appendix_background} for examples and further explanation.

Our human tactic proof step dataset consists of source-target pairs of
strings, one for each tactic invocation in the Lean core library and in
\texttt{mathlib}. The source string is the pretty-printed tactic
state. The target string is the tactic invocation as entered by a
human author of the source code.  
This data is gathered by hooking into the
Lean parser and Lean's compilation process. We refer to the task of
predicting the next human tactic proof step given a tactic state as
the {\em proofstep objective}.

\subsection{Proof artifact co-training} \label{subsec:pact}

In this section, we describe the PACT task suite and how data for these tasks are extracted.

For every proof term \(\tau\), we record the type
\(\Gamma\) of \(\tau\), its name
\texttt{nm}, and a list \texttt{ps} of all premises
(\ie named references to other lemmas in the library) which are used
in \(\tau\). We then recurse through \(\tau\), tracking a list \texttt{bs} of
bound variables which we update whenever navigating into the body of a
\(\lambda\)-expression. At every sub-term \(\tau' \subseteq \tau\) we record
\(\tau'\), its type \(\Gamma'\), the current state of \texttt{bs}, and the
following data:
\begin{enumerate}
\item A {\em tactic state}, where the goal is set to be
  \(\Gamma'\) and the list of
  hypotheses in the local context is set to be the list
  \texttt{bs}, \ie those bound variables in scope at \(\tau'\).
\item A {\em partial proof term}, \ie \(\tau\) with  \(\tau'\) masked
  out.
\item A {\em premise selection bitmask}, \ie Boolean labels for every
  \texttt{p} in \texttt{ps} indicating whether \texttt{p} is used in
  \(\tau'\).
\item A {\em local context bitmask}, \ie similar Boolean labels for every
  \texttt{b} in \texttt{bs} indicating whether \texttt{b} is used in
  \(\tau'\).
\item An optional {\em next lemma}: if the first step of \(\tau'\) is to apply a premise \texttt{p}
  in \texttt{ps}, we record \texttt{p}.
\end{enumerate}

Whenever we record a term, we record both \emph{pretty-printed}
and far more explicit \emph{fully elaborated} versions of it. The
fully elaborated terms
explicitly display enormous amounts
of type information which are usually
silently inferred by Lean. From these data, we assemble the following language modeling tasks:

\begin{enumerate}
\item {\bf Next lemma prediction.} Given the tactic state,
  predict the next lemma to be applied.

\item {\bf Proof term prediction.} Given the tactic state,
  predict the entire proof term \(\tau'\).

\item {\bf Skip-proof.} Given the partial proof term, predict the
  masked subterm \(\tau'\).
\item {\bf Type prediction.} Given the partial proof term, predict the
  type \(\Gamma'\) of the masked subterm \(\tau'\).

\item {\bf Tactic state elaboration.} Given the tactic state, predict
  the fully elaborated tactic state.

\item {\bf Proof term elaboration.} Given \(\tau\), predict the fully
  elaborated version of \(\tau\).

\item {\bf Premise classification.} Given the tactic state and a
  premise \texttt{p} \(\in\) \texttt{ps}, predict either
  \texttt{<TRUE>} or \texttt{<FALSE>} according to the premise
  selection bitmask.

\item {\bf Local context classification.} Given the tactic state
  (which consists of a list of local assumptions \texttt{bs} and the
  goal \(\Gamma'\)),
  predict the sublist of \texttt{bs} which is true on the local
  context bitmask.

\item {\bf Theorem naming.} Given the type \(\Gamma\) of the
  top-level proof term \(\tau\), predict the name \texttt{nm}.
\end{enumerate}



We remark that our next lemma prediction task is precisely
  the low-level \texttt{PROOFSTEP} objective studied
  in~\citep{DBLP:journals/corr/abs-2009-03393}, and our skip-proof
  task superficially resembles, but is much more difficult than the
  skip-tree task studied in~\citep{DBLP:journals/corr/abs-2006-04757},
  as proof terms tend to be far more complex than the syntax trees
  of theorem statements. 
\subsection{The {\sc LeanStep} machine learning environment} \label{subsec:proof-search}

We instrument Lean for
automatic theorem proving with a language
model, including utilities for (1) setting the runtime environment at
a particular theorem (ensuring proofs are
never circular), (2) serializing the tactic state as environment
observations for a theorem-proving agent, (3) exposing Lean's parser
to re-parse strings emitted by a language model into tactic
invocations, and (4) executing and capturing the results of the
re-parsed tactics, enabling the recording of trajectories for expert
iteration and reinforcement learning.

In addition to this general instrumentation, we implement a generic
best-first search algorithm for theorem proving; it forms
the basis for our evaluations and is written entirely in Lean itself.
The algorithm is
parametrized by an oracle \lstinline{(Ω : tactic_state → list (string × float))}
that accepts a tactic state and returns a list of strings and
heuristic scores. The search is controlled by a priority queue of {\em
search nodes}, which consist of a tactic state (\ie a partial proof)
and search metadata. In the outer loop of the algorithm---which
continues until either the theorem is completely proved (\ie no goals
are remaining on the current node), the priority queue is empty
(\ie the search has failed), or a pre-set timeout or budget of
iterations is exceeded---we pop a node off the queue,
serialize the associated tactic state and use
it to query the oracle, producing a list of candidates
\lstinline{cs : list (string × float)}. We then loop over the
candidates \lstinline{cs} to produce a list of new search nodes, by
re-parsing  each string into a tactic and adding a new node if the
parsed tactic advances the proof without raising errors.
These new search nodes are
then re-inserted into the queue in order of decreasing priority and
the search continues. We optionally constrain the search by enforcing
maximum width and depth limits \(w_{\operatorname{max}}\) and
\(d_{\operatorname{max}}\) that guard insertion into the queue. When
considering nodes for insertion, any node whose depth exceeds
\(d_{\operatorname{max}}\) is ignored, and all nodes are ignored if
the queue size is strictly larger than \(w_{\operatorname{max}}\). Due
to the flexibility in assigning heuristic scores and in choosing the
maximum width and depth hyperparameters, our algorithm is quite
general---for example, it reduces to (1) a greedy
depth-first search when \(w_{\operatorname{max}} = 0\), and (2) a
na{\"i}ve breadth-first search when heuristic scores are identical and
\(w_{\operatorname{max}} = d_{\operatorname{max}} = \infty\).



\section{Experiments}\label{sec:experiments}
\subparagraph{Training} \label{subsec:training}

In all of our experiments, we use decoder-only Transformers similar to
GPT-3~\citep{DBLP:conf/nips/BrownMRSKDNSSAA20}. Unless mentioned otherwise, all
of our models have \(24\) layers with \(d_{\operatorname{model}}=1536\) and
\(24\) heads, accruing to \(837\)M trainable parameters. They are also
pre-trained on \texttt{WebMath}~\citep{DBLP:journals/corr/abs-2009-03393}
for \(72\)B tokens. We use
the standard \texttt{BPE} encoding~\citep{DBLP:conf/nips/BrownMRSKDNSSAA20},
a batch size of \(512\) and a learning rate of \(0.00025\) with a cosine
schedule and a \(100\)-step ramp-up.

We use an 80-5-15
train-validation-test split. We split all datapoints deterministically
by {\em theorem name}, by hashing each name to a float in
\((0,1)\). This ensures, for example, that proof steps used to prove a
test theorem never appear in the training data and vice-versa.

When fine-tuning a model we load its saved parameters but re-initialize
the optimizer. We  
start each training for a fixed number of tokens (defining the cosine schedule)
and record the number of tokens consumed as we reach a minimal validation loss.
We use the minimum validation loss snapshot to evaluate each model on our
held-out test set.

We partition our datasets into three groups:
\begin{enumerate}
  \item \texttt{tactic}: the dataset described in \Cref{subsec:tactic}.
  \item \texttt{mix1}: the union of the PACT tasks {\bf next lemma
      prediction} and {\bf proof term prediction}~(\Cref{subsec:pact}),
    selected because of their close relation to \texttt{tactic}.
  \item \texttt{mix2}: all other datasets described in
    \Cref{subsec:pact}.
\end{enumerate}

This grouping is motivated by the impossibility to ablate each dataset
separately given our compute budget. They nonetheless enable us to study the
effect of tasks that are very close to the \texttt{tactic} objective in
comparison to others. Our choice of {\bf next lemma prediction} and {\bf proof
term prediction} for \texttt{mix1} is motivated by the observation that these
tasks are closely related to the theorem proving objective: a proof
can be given entirely in terms of a sequence of lemmas to apply (as in
Metamath), or the proof can be finished in one step by supplying the
entire proof term.
Despite their logical similarity to the \texttt{PROOFSTEP} objective,
we nevertheless use different keywords in the prompt to the
model to disambiguate (\texttt{NEXTLEMMA} and
\texttt{PROOFTERM}) from (\texttt{PROOFSTEP}) because
the data is noisy and represents a significant distribution shift: during
pretty-printing, subtrees of proof terms beyond a certain depth are dropped
entirely, there is generally no guarantee that they can be re-parsed, and the data
is much more verbose than what humans typically supply in source code.

\begin{figure*}
  \begin{center}
    \begin{tabular}{ll}
      \toprule
      \multicolumn{2}{l}{\texttt{tactic}} \\
      {\bf tactic proof steps} & \texttt{GOAL <TacticState> PROOFSTEP <Tactic>} \\
      \cmidrule(r){1-2}
      \multicolumn{2}{l}{\texttt{mix1}} \\
      {\bf next lemma prediction} & \texttt{GOAL <TacticState> NEXTLEMMA apply (<NextLemma>)} \\
      {\bf proof term prediction} & \texttt{GOAL <TacticState> PROOFTERM exact (<ProofTerm>)} \\
      \cmidrule(r){1-2}
      \multicolumn{2}{l}{\texttt{mix2}} \\
      {\bf skip proof} & \texttt{RESULT <MaskedProofTerm> SKIPPROOF <ProofTerm>} \\
      {\bf type prediction} & \texttt{RESULT <MaskedProofTerm> PREDICTTYPE <Type>} \\
      {\bf tactic state elaboration} & \texttt{GOAL <TacticState> ELABGOAL <ElaboratedTacticState>} \\
      {\bf proof term elaboration} & \texttt{PROOFTERM <ProofTerm> ELABPROOFTERM <ElaboratedProofTerm>} \\
      {\bf premise classification} & \texttt{GOAL <TacticState>
                                     CLASSIFYPREMISE <Premise> <True|False>} \\
      {\bf local context classification} & \texttt{GOAL <TacticState> CLASSIFYLOCALS <LocalsList>} \\
      {\bf theorem naming} & \texttt{TYPE <Type> NAME <Name>} \\
      \bottomrule
    \end{tabular}
    \caption{Auto-regressive objectives used for each task described in
    \Cref{sec:embed}. Placeholders represented with brackets (such as
    \texttt{<TacticState>}) are substituted by the context-completion pairs
    from each datasets in the prompts above. Each task is presented to the
    model with its respective keyword (\texttt{PROOFSTEP},
    \texttt{NEXTLEMMA},...). We wrap the completions of \texttt{mix1}
    tasks (with \texttt{apply(...)} and \texttt{exact(...)} respectively) as
    a hint that they are related to the respective Lean tactics; this is not
    directly possible for the other tasks.
  }
    \label{fig:training-objectives}
  \end{center}
\end{figure*}

\subparagraph{Theorem proving evaluation}

We run theorem-proving evaluations on our held-out \texttt{test} set,
comprising \(3071\) theorems.
Since the split was conducted by
theorem name, the proofs of these theorems never appear in the training data.
For each theorem in the test set, we set the runtime environment to the
location where the theorem is proved in the source code, preventing the use of
theorems defined later in \texttt{mathlib} and ensuring that we never
derive circular proofs.
We compare against existing proof automation In Lean by also
evaluating the tactics \texttt{refl}, which attempts to prove statements via
definitional equality, and \texttt{tidy}, which conducts a greedy
depth-first search using a fixed list of tactics at each step. We
re-implement \texttt{tidy} as a special case of our best-first search
algorithm using an oracle which always emits the same list of tactics,
and so henceforth refer to it as \texttt{tidy-bfs}.
In all of our experiments, we
use a maximum width of \(w_{\operatorname{max}}= 16\), a maximum depth
of \(d_{\operatorname{max}}=128\), a maximum budget of \(512\) iterations of the
outer loop, a timeout of \(5\) seconds per tactic execution, and a global
timeout of \(600\) seconds per theorem. Because sampling completions from our
models is much slower (\(\approx 1\) second) than
querying the constant \texttt{tidy-bfs} oracle (instantaneous), the
\texttt{tidy-bfs} search runs many more iterations
than \texttt{gptf} before timeout.

We report the pass-rate (\ie percentage of theorems proved) from the randomly-chosen
held-out test set, following
~\citep{DBLP:journals/corr/Whalen16}, \citep{bansal2019holist}, and others.
We provide an alternative pass-rate at the end of this section, using theorems
added to \texttt{mathlib} after our dataset was collected.
We average over three evaluation runs when reporting the pass rate.

\subparagraph{Effect of co-training vs pre-training}

We first study the effects of
pre-training versus co-training with the \texttt{mix1} and \texttt{mix2} datasets.
We pre-train using the methodology described above (potentially
pre-training first on \texttt{WebMath}, and then on a PACT dataset in sequence).
For co-training, we simply concatenate and shuffle the datasets together
without applying any particular weight to a given dataset.

The main results are presented in \Cref{fig:mainresults}. Pre-training
exhibits an effective transfer from \texttt{mix-1} and/or \texttt{mix-2} but
the best result is achieved by co-training with both these datasets. 
With this setup, we are able to train for much longer
(71B tokens vs 22B+18B for the best
pre-training setup) before overfitting on the \texttt{PROOFSTEP} task.
We hypothesize that PACT regularizes overfitting to the
\texttt{PROOFSTEP} task while still imparting useful knowledge to the
model due to large amounts of mutual information, 
and that this is the main driver of increased performance.

\begin{figure}[t]
  \begin{center}
    \begin{tabular}{lrrrrr}
      \toprule
      & Tokens & & & & \\
      Model & elapsed & \texttt{mix1} & \texttt{mix2} & \texttt{tactic} & Pass-rate\\
      \cmidrule(r){1-6}
      \multicolumn{5}{l}{\textit{Baselines}} \\
      \texttt{refl} & & & & & 1.1\% \\
      \texttt{tidy-bfs} & & & & & 9.9\% \\
      \texttt{WebMath > tactic} & 1B & & & 1.02 & 32.2\% \\
      \cmidrule(r){1-6}
      \multicolumn{5}{l}{\textit{Pre-training}} \\
      \texttt{WebMath > mix1} & 11B & \textit{0.08} & & & \\
      \texttt{WebMath > mix2} & 16B & & \textit{0.08} & & \\
      \texttt{WebMath > mix1 + mix2} & 22B & \textit{0.11} & \textit{0.08} & & \\
      \texttt{WebMath > mix1 > tactic} & 1B & & & 1.00 & 39.8\% \\
      \texttt{WebMath > mix1 + mix2 > tactic} & 1B & & & 0.97 & 44.0\% \\
      \cmidrule(r){1-6}
      \multicolumn{5}{l}{\textit{Co-training} (PACT)} \\
      \texttt{WebMath > mix1 + tactic} & 18B & \textit{0.08} & & 0.94 & 40.0\% \\
      \texttt{WebMath > mix2 + tactic} & 75B & & \textit{0.09} & 0.93 & 46.0\% \\
      \texttt{WebMath > mix1 + mix2 + tactic} & 71B & \textit{0.09} & \textit{0.09} & {\bf 0.91} & {\bf 48.4}\% \\
      \cmidrule(r){1-6}
      \multicolumn{5}{l}{\textit{Pre-training and co-training}} \\
      \texttt{WebMath > mix2 > mix1 + tactic} & 18B & \textit{0.08} & & 0.93 & 46.9\% \\
      \bottomrule
    \end{tabular}
    \caption{Comparison of pre-training and co-training on \texttt{mix-1} and
    \texttt{mix-2}. \texttt{>} denotes a pre-training step and \texttt{+}
    denotes a co-training. As an example, \texttt{WebMath > mix2 > mix1 +
    tactic} signifies a model successively pre-trained on \texttt{WebMath} then
    \texttt{mix2} and finally co-trained as a fine-tuning step on \texttt{mix1}
    and \texttt{tactic}. Columns \texttt{mix1}, \texttt{mix2}, \texttt{tactic}
    report the optimal validation loss achieved on these respective
    datasets. We provide a detailed description of experiment runtime
    and computing infrastructure in \Cref{sec:appendix_dataset}.}
    \label{fig:mainresults}
  \end{center}
\end{figure}

\subparagraph{Ablating \texttt{WebMath} pre-training}

Next, we ablate the effect of \texttt{WebMath} pre-training
(instead starting with a model pre-trained on the
same English language mix as GPT-3).  As
expected, co-trained models suffer from a performance drop without
\texttt{Webmath} pretraining. but we were more
interested in measuring the effect on pre-trained models on \texttt{mix-1} and
\texttt{mix-2}, as they may not benefit from \texttt{WebMath} as much due to
the two successive pre-training steps.

We report the optimal validation losses in ~\Cref{fig:webmathablate}.
\texttt{WebMath} appears as substantially beneficial even in the sequential
pre-training setup. This indicates that PACT is not a replacement for
\texttt{WebMath} pre-training, but rather a complementary method for
enhancing the performance of language models for theorem proving.


\subparagraph{Ablating regularization} We rule out the
possibility that the benefits from PACT come from simply regularizing
our models on the scarce \texttt{tactic} data alone. We checked that a \texttt{WebMath > tactic} model trained with
15\% residual dropout achieved
a minimum validation loss of 1.01 and 33.6\% pass rate, far below the
48.4\% PACT pass rate.

\begin{figure}[t]
  \begin{center}
    \begin{tabular}{lrrrrrr}
      \toprule
      & Tokens & Tokens & & & & \\
      Model & budget & elapsed & \texttt{mix1} & \texttt{mix2} & \texttt{tactic} & Pass-rate\textsuperscript{$\dagger$}\\
      \cmidrule(r){1-7}
      \multicolumn{5}{l}{\textit{Baselines}} \\
      \texttt{tactic} & 32B & 1B & & & 1.59 & --- \\
      \cmidrule(r){1-7}
      \multicolumn{5}{l}{\textit{Pre-training}} \\
      \texttt{mix1} & 32B & 20B & \textit{0.12} & & & \\
      \texttt{mix2} & 32B & 25B & & \textit{0.10} & & \\
      \texttt{mix1 + mix2} & 32B & 27B & \textit{0.13} & \textit{0.10} & & \\
      \texttt{mix1 > tactic} & 32B & 1B & & & 1.26 & --- \\
      \texttt{mix1 + mix2 > tactic} & 32B & 1B & & & 1.16 & --- \\
      \cmidrule(r){1-7}
      \multicolumn{5}{l}{\textit{Co-training}} \\
      \texttt{mix1 + tactic} & 32B & 27B & \textit{0.11} & & 1.12 & --- \\
      \texttt{mix2 + tactic} & 96B & 75B & &  \textit{0.10} & {\bf 1.02} &  40.4\% \\
      \texttt{mix1 + mix2 + tactic} & 96B & 71B & \textit{0.10} & \textit{0.11} & 1.07 & --- \\
      \cmidrule(r){1-7}
      \multicolumn{5}{l}{\textit{Pre-training and co-training}} \\
      \texttt{mix2 > mix1 + tactic} & 32B & 26B & \textit{0.11} & & 1.09 & --- \\
      \bottomrule
    \end{tabular}
    \caption{Validation losses achieved in the pre-training and co-training
    setups without \texttt{WebMath} pre-training. See
    \Cref{fig:mainresults} for a description of the columns and the
    models nomenclature used.
    \textsuperscript{$\dagger$}Due to technical
    constraints, we are unable to provide pass-rates for some of the
    models.%
    }
    \label{fig:webmathablate}
  \end{center}
\end{figure}

\subparagraph{Effect of model size}

Finally, we study how performance scales with respect to model
size. We use the best training setup
reported in \Cref{fig:mainresults}, \texttt{WebMath > mix1 + mix2 +
tactic}. The \texttt{837m} model is our main model. The \texttt{163m}
and \texttt{121m} models respectively have \(12\) and \(6\) layers, with
\(d_{\operatorname{model}}=768\). The learning rates are respectively adjusted
to \(0.0014\) and \(0.0016\).

As demonstrated by \Cref{fig:modelsize}, performance is highly correlated
with model size, with larger models generally achieving better generalization even
in the overfitted regime. We leave as future work a careful study of
how evaluation performance is affected when scaling to multi-billion
parameter models, as well as the feasibility of deploying them for
interactive use by Lean users.


\begin{figure}[t]
  \begin{center}
    \begin{tabular}{lrrrrrr}
      \toprule
      & Tokens & Tokens & & & & \\
      Model & budget & elapsed & \texttt{mix1} &
                                                          \texttt{mix2}
      & \texttt{tactic} & Pass-rate \\
      \cmidrule(r){1-7}
      \texttt{121m} & 96B & 82B & \textit{0.13} & \textit{0.10} & 1.23 & 35.1\% \\
      \texttt{163m} & 96B & 80B & \textit{0.12} & \textit{0.09} & 1.11 & 39.8\% \\
      \texttt{837m} & 96B & 71B & \textit{0.09} & \textit{0.09} & {\bf 0.91} & {\bf 48.4}\% \\
      \bottomrule
    \end{tabular}
    \caption{Validation losses and pass-rates achieved for various
      model sizes using PACT.
    See \Cref{fig:mainresults} for a description of the columns. 
 The setup used is \texttt{WebMath > mix1 + mix2 + tactic}.
  }
    \label{fig:modelsize}
  \end{center}
\end{figure}

\subparagraph{Time-stratified evaluation}

In the 5 week period that separated our last dataset extraction and the
writing of this paper, \texttt{mathlib} grew by 30K lines of code,
adding 2807 new theorems. Evaluating our models on these new theorem
statements gives a unique way to assess their capability to assist humans in
formalizing proofs and to test their generalization to completely unseen
theorems and definitions. This evaluation set also addresses one of
the weaknesses of using a random split of theorems from a formal
mathematics library, namely that the split is non-chronological; \eg
test theorems can appear as lemmas in proofs of train theorems.

We call this temporally held-out test set \texttt{future-mathlib} and evaluate
our best model as well as the \texttt{refl} and \texttt{tidy-bfs} baselines on
it. In contrast to evaluation on our \texttt{test} split, the
\texttt{refl} baseline (simply attempting a proof by the \texttt{refl}
tactic) closes \(328\) proofs (\(11.6 \%\)), demonstrating an important skew
towards trivial boilerplate lemmas generally defined to provide alternate
interfaces to new definitions. The \texttt{tidy-bfs} baseline closed \(611\)
proofs (\(21.8 \%\)), and our best model \texttt{wm-tt-m1-m2} closed \(1043\) proofs
(\(37.1 \%\)), proving \(94\%\) of the \texttt{refl} lemmas.
We attribute the weaker performance to 
heavy distribution shift: by the nature of the dataset,
the \texttt{future-mathlib} theorems frequently involve
new definitions and concepts which the model was never exposed to
during training. Nevertheless, the success rate remains high enough
to suggest strong generalization and usefulness at the frontier of
formalized mathematics.

\section{Discussion}\label{sec:discussion}
\subparagraph{Chained tactic predictions}\label{para:semicolon}
In Lean, multiple tactic commands can be chained together using
semicolons. Our data pipeline treats these tactic chains as a single
sequence in our training data, and they are occasionally predicted by
the model. Such chained tactic applications are difficult for human
formalizers to synthesize on their own, as they require reasoning
about the semantics of multiple tactics in sequence and their effects
on the tactic state, and the examples present in the training data are
usually optimized by hand from longer, less succinct proofs. We
observed that PACT significantly boosts the capability of our models
to {\em successfully} predict longer chained tactic applications. This
occurs despite the fact that the tactic chaining idiom is
specific to the tactic proofstep dataset and does not appear in the
PACT data whatsoever. We supply more detail in \Cref{subsec:chained_tactic_prediction}.

\subparagraph{Theorem naming} We also evaluate our best PACT
model (\texttt{wm-to-tt-m1-m2}) on the theorem naming task, using the
theorem statements and human-supplied names from the
\texttt{future-mathlib} evaluation set. It achieved 20\% acc@1, 27\%
acc@10, and 30\% acc@16. An inspection of its outputs reveals that
even when its predictions diverge from the ground truth, they are
often idiomatic and semantically correct alternatives. We supply more
detail in \Cref{subsec:theorem_naming}.


\subparagraph{Impact on Lean community}

Lean's \texttt{mathlib}~\citep{DBLP:conf/cpp/X20} is a rapidly
growing open source library of formal mathematics which has grown
considerably in size each year for the past four years.\footnote{See
\url{https://leanprover-community.github.io/mathlib_stats.html}
for up-to-date statistics on \texttt{mathlib}'s size and growth over
time.} Our work has been welcomed by members
of this community, with Lean power users describing some of the new
proofs found by GPT-f as ``nontrivial'' and ``clever''.
More than one-third of the proofs found by our models are shorter and
produce smaller proof terms (sometimes by several orders of
magnitude) than the ground truth. Manually inspecting a small,
non-cherry picked sample of
these shorter proofs has led to 19 GPT-f co-authored commits to
\texttt{mathlib}, some of which reduce proof term sizes and theorem
compilation times by an order of magnitude (see \Cref{sec:appendix_examples}).

\subparagraph{Potential societal impact} \label{societal-impact}
Strong automated reasoning systems have enormous potential impact for
mathematical research and scientific progress in other
disciplines. The methods that we discuss in this paper could
accelerate the development of strong automated reasoning systems. We
have also observed that our language models absorb stylistic biases from their
training data which could be amplified via reinforcement learning.
However, since we focus on mathematics codified in proof
assistants, we believe that there is little immediate negative
societal impact from our work.


\subparagraph{Future directions}
There are many elaborations on the training data, training
methodology, and tree search wrapping \texttt{lean-gptf} which can
be reasonably expected to improve its performance at theorem proving.
Our dataset can be synthetically augmented using similar methods as \citep{DBLP:journals/corr/abs-2009-03393}.
Our dataset could be cleaned further, and proofs
minimized. Merely making the decoded rewrites robust by only using the
largest prefix of successful rewrites significantly boosts the success
rate of suggested rewrites. In a similar vein, predicted lemmas
generated as arguments to unsuccessful tactic
applications could be cached and re-used as hints for an
intermittently-queried hammer. The increased success rate of
chained tactic predictions mentioned above shows the
feasibility of having language models perform multiple reasoning steps
in a single query, potentially improving the efficiency
of the proof search. From the experiments described in
\Cref{sec:experiments}, it is clear that the composition of the
dataset used for co-training significantly affects performance on
theorem proving. Although we uniformly sampled across all co-training
tasks, it would be interesting to optimize a dynamic mixture schedule,
perhaps annealing towards a desired task.


\subparagraph{Conclusion}

There is a sense in which PACT is merely an application of the well
known principle that compute in the form of search should be exchanged
for training signal whenever possible. In Lean, typeclass inference
relies on a backtracking Prolog-style search; the elaborator performs
search to disambiguate overloaded notation and infer types; Lean tactics
have complex semantics precisely because they can perform
search to find subproofs automatically. The work done by these
subroutines is preserved in the proof artifacts,
and PACT can be viewed as a way of extracting this information offline
for more training signal.

We have presented PACT as a way of addressing the data scarcity issue
for learning theorem proving from human tactic scripts in proof
assistant libraries. Another well-studied solution for this is
expert iteration and reinforcement learning. In the setting of HOL
Light, and under the assumption of a hardcoded finite action space of
tactics, \cite{bansal2019learning} in conjunction with supervised seed data
was able to achieve up to 70\% proof success rate on the HOList
theorem proving task. Similarly, in a set-up much closer to ours, MM
GPT-f demonstrated the feasibility of expert iteration when using
generative language models for theorem proving.

Within a fixed corpus of theorems (and hence proof terms), however,
both PACT and RL are fundamentally constrained by a lack of
exploration---as the performance of the theorem proving agent
improves, it will eventually saturate and become starved for data, and
its curriculum will need to be expanded. Although self-supervised methods
such as PACT represent a way to significantly improve the
data-efficiency of reinforcement learning loops over existing theorem
prover libraries, the development of continuously self-improving and
infinitely scalable neural theorem provers remains contingent on
sufficiently powerful exploration and automated curriculum generation;
we consider these challenges to be of paramount importance.

\section{Acknowledgments}\label{sec:acknowledgement}
We thank the members of the Lean community, in particular Kevin
Buzzard, Simon Hudon, Johan Commelin, Mario Carneiro, Bhavik Mehta,
and Gabriel Ebner for their valuable feedback on our work. We are
indebted to Markus Rabe and Christian Szegedy for many hours of
helpful discussion. We also thank Daniel Selsam, Tom Hales, and Josef
Urban for feedback on earlier drafts of this paper.

\section{Reproducibility statement}\label{sec:source_code}
The source code used to generate the Lean datasets and run the evaluation
is open source and made available in the following repositories:
\begin{description}
	\item[Lean theorem proving environment]: \\
	\url{https://github.com/jesse-michael-han/lean-tpe-public}
	\item[Tactic step data pipeline]: \\
	\url{https://github.com/jasonrute/lean_proof_recording}
	\item[PACT data pipeline]: \\
	\url{https://github.com/jesse-michael-han/lean-step-public}
\end{description}

Our Transformer model was pre-trained on two proprietary datasets.
The first is the same
mix used by GPT-3~\citep{DBLP:conf/nips/BrownMRSKDNSSAA20}
and the second is \texttt{WebMath}~\citep{DBLP:journals/corr/abs-2009-03393}.
More details can be found in \Cref{sec:appendix_dataset}.

While our weights and the API through which we query our models are not currently
public, techniques for training decoder-only transformers and efficiently performing
inference with them are well-known.
Our released theorem proving code is agnostic to these implementation details and
will work with any language model exposed via an HTTP server.
The provided code also supports querying a locally hosted Transformer from the
open-source library \texttt{fairseq} via the Fairseq
CLI~\citep{DBLP:conf/naacl/OttEBFGNGA19}.

We have released a simplified version of the proof search described in
\Cref{subsec:proof-search} as a tactic to the Lean community in a public beta,
opening the way for our models to directly accelerate the development of
formalized mathematics and for human experts to provide feedback and
additional training signal in a virtuous cycle.  The tactic and code are available
at \url{https://github.com/jesse-michael-han/lean-gptf},
and users who sign up for the beta are granted access
to our Transformer model through an API.

\bibliography{lean-step-iclr} \bibliographystyle{iclr2022_conference}

\newpage
\appendix

\section{Additional Background}\label{sec:appendix_background}
\subparagraph*{Proof terms}
Lean's fundamental logic is a dependent type theory called the
calculus of inductive constructions~\cite{DBLP:conf/mfps/PfenningP89}.
This design means that terms ($4$, $x + y$, $f$), types
($\mathbb{N}$, $\mathtt{list} \; \mathbb{Z}$, $\alpha \to \beta$)
and proofs are all represented with a single datatype called an
\emph{expression}.
Given an environment of available constants and definitions and
a context $\Gamma$ of variables, Lean can infer a type
$\alpha$ for each well-formed expression $t$.
A \emph{proof term} is a Lean expression whose type is
a proposition. This proof term serves as a checkable artifact for
verifying the proposition. Lean uses a small,
trusted kernel to verify proof terms.

\subparagraph*{Tactics}

Tactics in Lean are
metaprograms~\cite{DBLP:journals/pacmpl/EbnerURAM17}, which can
construct Lean expressions, such as terms. A
{\em tactic state} which tracks the list of open goals and other
metadata is threaded through each tactic invocation.
Lean has special support for treating tactics as an extensible
domain-specific language (DSL); this DSL is how Lean is typically used
as an interactive theorem prover. The DSL amounts to a linear chain of
comma-separated invocations. The process of interactive proving is
mediated through Lean's language server, which will present the
context and type for the current goal in the proof to the user, dependent on where their
cursor is in the source text. The {\em tactic prediction} task is to predict the next tactic given
this goal state. We extract supervised training data for this task by
extracting all human-supplied proof steps from Lean's \texttt{mathlib}.

An object called the \emph{tactic state} is threaded through each invocation of a tactic.
Among other things, the tactic state maintains a context of metavariables: placeholders in to which expressions will be substituted later.
At each point in the proof, one or more of these metavariables are selected as the \emph{goal} of the tactic state which is present
As the proof progresses, there are multiple values to be found

\subparagraph*{Example}

Consider this (modified) example of a tactic proof from the library.
\begin{lstlisting}
theorem int.sub_ne_zero_of_ne : ∀ (a b : ℤ), a ≠ b -> a - b ≠ 0 :=
begin
	intros a b h hab,
	apply h,
	apply int.eq_of_sub_eq_zero hab,
end
\end{lstlisting}

Each tactic line modifies the proof state, which we explicitly annotate
below with comments between each tactic.
\begin{lstlisting}
theorem int.sub_ne_zero_of_ne : ∀ (a b : ℤ), a ≠ b -> a - b ≠ 0 :=
begin
  -- ⊢ ∀ (a b : ℤ), a ≠ b → a - b ≠ 0
  intros a b h hab,
  -- a b : ℤ,
  -- h : a ≠ b,
  -- hab : a - b = 0
  -- ⊢ false
  apply h,
  -- a b : ℤ,
  -- h : a ≠ b,
  -- hab : a - b = 0
  -- ⊢ a = b
  apply int.eq_of_sub_eq_zero hab,
  -- no goals
end
\end{lstlisting}
Our proofstep objective is to predict the tactic applied to a given tactic state.

Lean stores this proof internally as a proof term:
\begin{lstlisting}
theorem int.sub_ne_zero_of_ne : ∀ (a b : ℤ), a ≠ b → a - b ≠ 0 :=
λ (a b : ℤ) (h : a ≠ b), id (λ (hab : a - b = 0), h (int.eq_of_sub_eq_zero hab))
\end{lstlisting}

Since this proof term is just stored internally as a tree,
any branch of this term tree can be removed,
to create a hole \verb|_|, for example:
\begin{lstlisting}
λ (a b : ℤ) (h : a ≠ b), id (λ (hab : a - b = 0), h  _)
\end{lstlisting}
Lean will automatically provide a list of both the local context and the type of
a term needed to fill that hole as shown below.
Notice this is the same as a tactic state we
saw from the term proof above.
\begin{lstlisting}
a b : ℤ,
h : a ≠ b,
hab : a - b = 0
⊢ a = b
\end{lstlisting}
Using this methodology of following proof term trees,
we can mine low level proof data for every node of a term proof
to produce the PACT dataset described in \Cref{subsec:pact}.

\section{Datasets}\label{sec:appendix_dataset}
\subsection{Pre-training datasets}

We pre-train on \texttt{WebMath} as described
in~\citep{DBLP:journals/corr/abs-2009-03393}.
All models, including the \texttt{WebMath} pre-trained
models, and the non-\texttt{WebMath} models used in ablations,
were first pre-trained on the mix used by
GPT-3~\citep{DBLP:conf/nips/BrownMRSKDNSSAA20} which includes a filtered
\texttt{CommonCrawl}, \texttt{WebText2}, \texttt{Book1}, \texttt{Book2} and
\texttt{Wikipedia}. \texttt{WebMath} includes Python-only \texttt{GitHub} data,
as well as \texttt{arXiv} and \texttt{Math StackExchange}.

From these datasets, a potential risk for test-set contamination (presence of
\texttt{mathlib}) exists for the crawled datasets, namely \texttt{CommonCrawl},
\texttt{WebText2}, and (in case of a filtering bug) Python-only
\texttt{GitHub}. The other datasets (in particular \texttt{arXiv} and
\texttt{Math StackExchange}) may contain short references of \texttt{mathlib}
code but in shape and forms that would not lead to effective contamination.

To assess the contamination risk related with the crawled datasets,
we searched
\texttt{CommonCrawl}, \texttt{WebText2}, \texttt{arXiv}, Python-only
\texttt{GitHub}, and \texttt{Math StackExchange} for test theorems.
For example, given the test theorem \verb|nat.div_eq_sub_div| we
searched for any occurrences of the string \verb|div_eq_sub_div|.
Of over 3000 test theorem names,
we found 595 which occurred in the datasets.
Many instances were innocuous, but some were in Lean
files, and in some cases there was a proof of a test theorem.
There were also 160 additional test theorems with no underscore in their name,
which we did not check, but whose name is likely to be found in the datasets.
(There is no need to check for training theorems since they are already in the
training data and it would not constitute contamination.)
We re-calculated the pass-rates of the results in \Cref{fig:mainresults}
omitting these 755 test theorems.
This decreases the reported pass-rates slightly, ranging from $0.6$ to
$1.1$ percentage points.
The adjusted pass-rate of our best model \texttt{WebMath > mix1 + mix2 + tactic}
is $47.4\%,$ a decrease of $1$ percentage point. Our
main results still hold even with the adjusted pass-rates.

Additionally we also look at the results for the 1,350 test theorems
in our dataset that were added
to Lean and \texttt{mathlib} after April 18, 2020, which is after
 \texttt{CommonCrawl} and \texttt{WebText2} were gathered,
and the 544 test theorems added after September 11, 2020, which is
after \texttt{WebMath} was gathered.
Unlike \verb|future-mathlib|, these theorems were part of the originally
extracted data.
The pass-rates for the \texttt{WebMath > mix1 + mix2 + tactic} model
on these restricted sets of test theorems are $45.6\%$ and $43.3\%$,
respectively.

We also looked for the following Metamath specific and HOL specific strings
in \texttt{CommonCrawl}, \texttt{WebText2}, and Python-only \texttt{GitHub}:
\begin{lstlisting}
Metamath:
    "( ph -> A = C )"
    "( ph -> A R C )"
    "( sqrt ` 2 ) e/ QQ"
HOL:
    "apply (rule "
    "apply (drule "
\end{lstlisting}

We found \(0\) occurrence of the Metamath-related strings but interestingly
found a non-negligible amount of HOL-related documents, which does not
constitute a test-set contamination but potentially benefits the downstream
tasks studied in this paper.

While our results show a significant benefit to pre-training on \texttt{WebMath},
it is unclear exactly how pre-training helps.
Since Lean's theorem names are made of coded mathematical phases,
e.g.\ \verb|affine.simplex.dist_circumcenter_eq_circumradius|,
it is not unreasonable to suspect that important statistical connections
are extracted from math sources.
It is even possible that simple instances of auto-formalization
or ITP translation are happening.
There is prior work \citep{DBLP:conf/lpar/GauthierK15,DBLP:conf/mkm/WangKU18,
DBLP:conf/cpp/WangBKU20}
suggesting that both of these are possible.
From the point of view of a
\verb|lean-gptf| end-user,
any such extraction of prior, publicly available data is useful and helpful.
Nonetheless, our results are of a different nature than other AI for
theorem proving research which do not use data outside of a given
theorem proving library.
This should be taken into account in any future comparisons and benchmarks.

\subsection{Dataset sizes}
\begin{itemize}
\item \texttt{tactic}: \(\approx\)\(128\mathrm{K}\) examples.
\item \texttt{mix1}
  \begin{itemize}
  \item {\bf Next lemma prediction}: \(\approx\)\(2.5\mathrm{M}\) examples
  \item {\bf Proof term prediction}: \(\approx\)\(2.9\mathrm{M}\) examples
  \end{itemize}
\item \texttt{mix2}
  \begin{itemize}
  \item {\bf Skip-proof}: \(\approx\)\(1.7\mathrm{M}\) examples
  \item {\bf Type-prediction}: \(\approx\)\(1.7\mathrm{M}\) examples
  \item {\bf Tactic state elaboration}: \(\approx\)\(346\mathrm{K}\)
    examples
  \item {\bf Proof term elaboration}: \(\approx\)\(1.0\mathrm{M}\) examples
  \item {\bf Premise classification}: \(\approx\)\(9.3\mathrm{M}\) examples
  \item {\bf Local context classification}: \(\approx\)\(2.0\mathrm{M}\)
    examples
  \item {\bf Theorem naming}: \(\approx\)\(32K\) examples.
  \end{itemize}
\end{itemize}
\subsection{Example datapoints}

We present datapoints extracted from a toy example, namely the proof of the Peirce
identity, viz.

\begin{lstlisting}
lemma peirce_identity {P Q :Prop} : ((P → Q) → P) → P :=
begin
  apply or.elim (em P),
  intros h _,
  exact h,
  tauto!
end
\end{lstlisting}

From this, we can extract four \texttt{tactic} datapoints (i.e. human-generated
tactic proof steps):

\begin{lstlisting}
-- GOAL P Q : Prop ⊢ ((P → Q) → P) → P PROOFSTEP apply or.elim (em P)
-- GOAL P Q : Prop ⊢ P → ((P → Q) → P) → P  P Q : Prop ⊢ ¬P → ((P → Q) → P) → P PROOFSTEP intros h _
-- GOAL P Q : Prop, h : P, ᾰ : (P → Q) → P ⊢ P  P Q : Prop ⊢ ¬P → ((P → Q) → P) → P PROOFSTEP exact h
-- GOAL P Q : Prop ⊢ ¬P → ((P → Q) → P) → P PROOFSTEP tauto!
\end{lstlisting}

In contrast, we can extract dozens of raw PACT datapoints. Due to
space constraints, we list a representative sample of four such datapoints,
from each of which we can derive the nine self-supervised auxiliary PACT tasks studied in our
present work. For example, proof term prediction
is precisely predicting the \lstinline{"proof_term"} given the
concatenation of \lstinline{"hyps"}, \lstinline{"⊢"}, and the
\lstinline{"goal"}, skip-proof is predicting the
\lstinline{"proof_term"} given \lstinline{"result"}, etc.

\begin{lstlisting}
  DATAPOINT:
---
{ "decl_nm":"peirce_identity",
  "decl_tp":"∀ {P Q : Prop}, ((P → Q) → P) → P",
  "hyps":[["P", "Prop"], ["Q", "Prop"], ["ᾰ", "¬P"], ["ᾰ_1", "(P → Q) → P"], ["ᾰ_1", "¬(P → Q)"]],
  "hyps_mask":[true, false, false, false, false],
  "decl_premises":[["absurd", "∀ {a b : Prop}, a → ¬a → b"],
   ["absurd", "∀ {a b : Prop}, a → ¬a → b"],
   ["decidable.not_imp", "∀ {a b : Prop} [_inst_1 : decidable a], ¬(a → b) ↔ a ∧ ¬b"],
   ["iff.mp", "∀ {a b : Prop}, (a ↔ b) → a → b"],
   ["and.dcases_on",
    "∀ {a b : Prop} {C : a ∧ b → Prop} (n : a ∧ b), (∀ (left : a) (right : b), C _) → C n"],
   ["decidable.not_or_of_imp", "∀ {a b : Prop} [_inst_1 : decidable a], (a → b) → ¬a ∨ b"],
   ["or.dcases_on",
    "∀ {a b : Prop} {C : a ∨ b → Prop} (n : a ∨ b), (∀ (h : a), C _) → (∀ (h : b), C _) → C n"],
   ["em", "∀ (p : Prop), p ∨ ¬p"],
   ["or.elim", "∀ {a b c : Prop}, a ∨ b → (a → c) → (b → c) → c"]],
  "decl_premises_mask":[false, false, true, false, false, false, false, false, false],
  "goal":"∀ {b : Prop} [_inst_1 : decidable P], ¬(P → b) ↔ P ∧ ¬b",
  "proof_term":"decidable.not_imp",
  "result":"λ {P Q : Prop}, (em P).elim (λ (h : P) (ᾰ : (P → Q) → P), h) (λ (ᾰ : ¬P) (ᾰ_1 : (P → Q) → P), (decidable.not_or_of_imp ᾰ_1).dcases_on (λ (ᾰ_1 : ¬(P → Q)), ((PREDICT Q (classical.prop_decidable P)).mp ᾰ_1).dcases_on (λ (ᾰ_1_left : P) (ᾰ_1_right : ¬Q), absurd ᾰ_1_left ᾰ)) (λ (ᾰ_1 : P), absurd ᾰ_1 ᾰ))",
  "next_lemma":["decidable.not_imp", "∀ {a b : Prop} [_inst_1 : decidable a], ¬(a → b) ↔ a ∧ ¬b"],
  "goal_is_prop":true,
  "verbose_proof_term":"@decidable.not_imp P",
  "verbose_goal":"∀ {b : Prop} [_inst_1 : decidable P], ¬(P → b) ↔ P ∧ ¬b",
  "verbose_result":"λ {P Q : Prop}, (em P).elim (λ (h : P) (ᾰ : (P → Q) → P), h) (λ (ᾰ : ¬P) (ᾰ_1 : (P → Q) → P), (@decidable.not_or_of_imp (P → Q) P (classical.prop_decidable (P → Q)) ᾰ_1).dcases_on (λ (ᾰ_1 : ¬(P → Q)), (@iff.mp (¬(P → Q)) (P ∧ ¬Q) (PREDICT Q (classical.prop_decidable P)) ᾰ_1).dcases_on (λ (ᾰ_1_left : P) (ᾰ_1_right : ¬Q), @absurd P P ᾰ_1_left ᾰ)) (λ (ᾰ_1 : P), @absurd P P ᾰ_1 ᾰ))"}
---

DATAPOINT:
---
{ "decl_nm":"peirce_identity",
  "decl_tp":"∀ {P Q : Prop}, ((P → Q) → P) → P",
  "hyps":[["P", "Prop"], ["Q", "Prop"], ["ᾰ", "¬P"], ["ᾰ_1", "(P → Q) → P"], ["ᾰ_1", "¬(P → Q)"]],
  "hyps_mask":[false, true, false, false, false],
  "decl_premises":[["absurd", "∀ {a b : Prop}, a → ¬a → b"],
   ["absurd", "∀ {a b : Prop}, a → ¬a → b"],
   ["decidable.not_imp", "∀ {a b : Prop} [_inst_1 : decidable a], ¬(a → b) ↔ a ∧ ¬b"],
   ["iff.mp", "∀ {a b : Prop}, (a ↔ b) → a → b"],
   ["and.dcases_on",
    "∀ {a b : Prop} {C : a ∧ b → Prop} (n : a ∧ b), (∀ (left : a) (right : b), C _) → C n"],
   ["decidable.not_or_of_imp", "∀ {a b : Prop} [_inst_1 : decidable a], (a → b) → ¬a ∨ b"],
   ["or.dcases_on",
    "∀ {a b : Prop} {C : a ∨ b → Prop} (n : a ∨ b), (∀ (h : a), C _) → (∀ (h : b), C _) → C n"],
   ["em", "∀ (p : Prop), p ∨ ¬p"],
   ["or.elim", "∀ {a b c : Prop}, a ∨ b → (a → c) → (b → c) → c"]],
  "decl_premises_mask":[false, false, false, false, false, false, false, false, false],
  "goal":"Prop",
  "proof_term":"Q",
  "result":"λ {P Q : Prop}, (em P).elim (λ (h : P) (ᾰ : (P → Q) → P), h) (λ (ᾰ : ¬P) (ᾰ_1 : (P → Q) → P), (decidable.not_or_of_imp ᾰ_1).dcases_on (λ (ᾰ_1 : ¬(P → Q)), (decidable.not_imp.mp ᾰ_1).dcases_on (λ (ᾰ_1_left : P) (ᾰ_1_right : ¬Q), absurd ᾰ_1_left ᾰ)) (λ (ᾰ_1 : P), absurd ᾰ_1 ᾰ))",
  "next_lemma":["Q", "Prop"],
  "goal_is_prop":false,
  "verbose_proof_term":"Q",
  "verbose_goal":"Prop",
  "verbose_result":"λ {P Q : Prop}, (em P).elim (λ (h : P) (ᾰ : (P → Q) → P), h) (λ (ᾰ : ¬P) (ᾰ_1 : (P → Q) → P), (@decidable.not_or_of_imp (P → Q) P (classical.prop_decidable (P → Q)) ᾰ_1).dcases_on (λ (ᾰ_1 : ¬(P → Q)), ((@decidable.not_imp P PREDICT (classical.prop_decidable P)).mp ᾰ_1).dcases_on (λ (ᾰ_1_left : P) (ᾰ_1_right : ¬Q), @absurd P P ᾰ_1_left ᾰ)) (λ (ᾰ_1 : P), @absurd P P ᾰ_1 ᾰ))"}
---

DATAPOINT:
---
{ "decl_nm":"peirce_identity",
  "decl_tp":"∀ {P Q : Prop}, ((P → Q) → P) → P",
  "hyps":[["P", "Prop"], ["Q", "Prop"], ["ᾰ", "¬P"], ["ᾰ_1", "(P → Q) → P"], ["ᾰ_1", "¬(P → Q)"]],
  "hyps_mask":[true, true, false, false, false],
  "decl_premises":[["absurd", "∀ {a b : Prop}, a → ¬a → b"],
   ["absurd", "∀ {a b : Prop}, a → ¬a → b"],
   ["decidable.not_imp", "∀ {a b : Prop} [_inst_1 : decidable a], ¬(a → b) ↔ a ∧ ¬b"],
   ["iff.mp", "∀ {a b : Prop}, (a ↔ b) → a → b"],
   ["and.dcases_on",
    "∀ {a b : Prop} {C : a ∧ b → Prop} (n : a ∧ b), (∀ (left : a) (right : b), C _) → C n"],
   ["decidable.not_or_of_imp", "∀ {a b : Prop} [_inst_1 : decidable a], (a → b) → ¬a ∨ b"],
   ["or.dcases_on",
    "∀ {a b : Prop} {C : a ∨ b → Prop} (n : a ∨ b), (∀ (h : a), C _) → (∀ (h : b), C _) → C n"],
   ["em", "∀ (p : Prop), p ∨ ¬p"],
   ["or.elim", "∀ {a b c : Prop}, a ∨ b → (a → c) → (b → c) → c"]],
  "decl_premises_mask":[false, false, true, false, false, false, false, false, false],
  "goal":"∀ [_inst_1 : decidable P], ¬(P → Q) ↔ P ∧ ¬Q",
  "proof_term":"decidable.not_imp",
  "result":"λ {P Q : Prop}, (em P).elim (λ (h : P) (ᾰ : (P → Q) → P), h) (λ (ᾰ : ¬P) (ᾰ_1 : (P → Q) → P), (decidable.not_or_of_imp ᾰ_1).dcases_on (λ (ᾰ_1 : ¬(P → Q)), ((PREDICT (classical.prop_decidable P)).mp ᾰ_1).dcases_on (λ (ᾰ_1_left : P) (ᾰ_1_right : ¬Q), absurd ᾰ_1_left ᾰ)) (λ (ᾰ_1 : P), absurd ᾰ_1 ᾰ))",
  "next_lemma":["decidable.not_imp", "∀ {a b : Prop} [_inst_1 : decidable a], ¬(a → b) ↔ a ∧ ¬b"],
  "goal_is_prop":true,
  "verbose_proof_term":"@decidable.not_imp P Q",
  "verbose_goal":"∀ [_inst_1 : decidable P], ¬(P → Q) ↔ P ∧ ¬Q",
  "verbose_result":"λ {P Q : Prop}, (em P).elim (λ (h : P) (ᾰ : (P → Q) → P), h) (λ (ᾰ : ¬P) (ᾰ_1 : (P → Q) → P), (@decidable.not_or_of_imp (P → Q) P (classical.prop_decidable (P → Q)) ᾰ_1).dcases_on (λ (ᾰ_1 : ¬(P → Q)), (@iff.mp (¬(P → Q)) (P ∧ ¬Q) (PREDICT (classical.prop_decidable P)) ᾰ_1).dcases_on (λ (ᾰ_1_left : P) (ᾰ_1_right : ¬Q), @absurd P P ᾰ_1_left ᾰ)) (λ (ᾰ_1 : P), @absurd P P ᾰ_1 ᾰ))"}
---

DATAPOINT:
---
{ "decl_nm":"peirce_identity",
  "decl_tp":"∀ {P Q : Prop}, ((P → Q) → P) → P",
  "hyps":[["P", "Prop"], ["Q", "Prop"], ["ᾰ", "¬P"], ["ᾰ_1", "(P → Q) → P"], ["ᾰ_1", "¬(P → Q)"]],
  "hyps_mask":[false, false, false, false, false],
  "decl_premises":[["absurd", "∀ {a b : Prop}, a → ¬a → b"],
   ["absurd", "∀ {a b : Prop}, a → ¬a → b"],
   ["decidable.not_imp", "∀ {a b : Prop} [_inst_1 : decidable a], ¬(a → b) ↔ a ∧ ¬b"],
   ["iff.mp", "∀ {a b : Prop}, (a ↔ b) → a → b"],
   ["and.dcases_on",
    "∀ {a b : Prop} {C : a ∧ b → Prop} (n : a ∧ b), (∀ (left : a) (right : b), C _) → C n"],
   ["decidable.not_or_of_imp", "∀ {a b : Prop} [_inst_1 : decidable a], (a → b) → ¬a ∨ b"],
   ["or.dcases_on",
    "∀ {a b : Prop} {C : a ∨ b → Prop} (n : a ∨ b), (∀ (h : a), C _) → (∀ (h : b), C _) → C n"],
   ["em", "∀ (p : Prop), p ∨ ¬p"],
   ["or.elim", "∀ {a b c : Prop}, a ∨ b → (a → c) → (b → c) → c"]],
  "decl_premises_mask":[false, false, false, false, false, false, false, false, false],
  "goal":"Π (a : Prop), decidable a",
  "proof_term":"classical.prop_decidable",
  "result":"λ {P Q : Prop}, (em P).elim (λ (h : P) (ᾰ : (P → Q) → P), h) (λ (ᾰ : ¬P) (ᾰ_1 : (P → Q) → P), (decidable.not_or_of_imp ᾰ_1).dcases_on (λ (ᾰ_1 : ¬(P → Q)), (decidable.not_imp.mp ᾰ_1).dcases_on (λ (ᾰ_1_left : P) (ᾰ_1_right : ¬Q), absurd ᾰ_1_left ᾰ)) (λ (ᾰ_1 : P), absurd ᾰ_1 ᾰ))",
  "next_lemma":["classical.prop_decidable", "Π (a : Prop), decidable a"],
  "goal_is_prop":false,
  "verbose_proof_term":"classical.prop_decidable",
  "verbose_goal":"Π (a : Prop), decidable a",
  "verbose_result":"λ {P Q : Prop}, (em P).elim (λ (h : P) (ᾰ : (P → Q) → P), h) (λ (ᾰ : ¬P) (ᾰ_1 : (P → Q) → P), (@decidable.not_or_of_imp (P → Q) P (PREDICT (P → Q)) ᾰ_1).dcases_on (λ (ᾰ_1 : ¬(P → Q)), ((@decidable.not_imp P Q (PREDICT P)).mp ᾰ_1).dcases_on (λ (ᾰ_1_left : P) (ᾰ_1_right : ¬Q), @absurd P P ᾰ_1_left ᾰ)) (λ (ᾰ_1 : P), @absurd P P ᾰ_1 ᾰ))"}
---
\end{lstlisting}

\section{Experiments}\label{sec:appendix_experiments}
\subsection{Chained tactic prediction}\label{subsec:chained_tactic_prediction}

Individual Lean tactics are chained together with commas. However, the
Lean interactive tactic DSL also includes
a number of other tactic combinators for creating composite
tactics. A frequently used combinator is the infix semicolon
\verb`t; s` which will perform the tactic \verb`t` and then apply the
tactic \verb`s` to each of the resulting subgoals produced by
\verb`t`.
Our data pipeline for human tactic proof steps treats
these semicolon-chained tactics as a single string for the language
modeling objective. Thus, our models learn to occasionally emit
multiple-step tactic predictions using semicolons.
For example, \texttt{wm-to-tt-m1-m2} solved the following lemma in
category theory with a single prediction chaining four tactics in a row:

\begin{lstlisting}
  theorem category_theory.grothendieck.congr
    {X Y : grothendieck F} {f g : X ⟶ Y} (h : f = g) :
    f.fiber = eq_to_hom (by subst h) ≫ g.fiber :=
  begin
    rcases X; rcases Y; subst h; simp
  end
\end{lstlisting}

One way of measuring the sophistication of predicted tactics is to
consider the number of successful proofs on the evaluation
set which have this composite form using semicolon-chaining. We
display this analysis in \Cref{fig:semicolon-table}, which shows that
training
with PACT in addition to the human-made tactics causes longer
semicolon-chained tactics to be successfully predicted during theorem proving. This
is remarkable because the semicolon idiom is specific to the tactic
DSL which does not occur in the PACT data whatsoever,
and yet the co-training causes longer and more frequent
successful composite tactic predictions.

\begin{table}[b]
    \cprotect\caption{Counting the number of
semicolon-chained tactics predicted by our models that appear {\em
  in successful proofs}.  Each column headed by a number
\(n\mathtt{;}\) indicates the number of times that a suggestion
appeared with \(n\) occurrences of `\verb`;`'.}
    \label{fig:semicolon-table}
    \vskip 0.15in
    \begin{center}
    \begin{small}

    \begin{tabular}{lrrrrr}
    \hline

    {\sc Model}        & 1\verb`;` & 2\verb`;` & 3\verb`;` &
                                                                  4\verb`;`   & {\sc Mean} \\
    \hline

\texttt{wm-to-tt}           & 215         & 49 & 2  & 0 & 1.199 \\
\texttt{wm-to-tt-m1}      & 186         & 39 & 5  & 1 & 1.225 \\

      \texttt{wm-to-tt-m1-m2} & {\bf 328}         & {\bf 82} & {\bf 12}
                                                           & {\bf 3} & {\bf 1.271} \\
    \hline
    \end{tabular}

    \end{small}
    \end{center}
    \vskip -0.1in
\end{table}

\subsection{Theorem naming case study}\label{subsec:theorem_naming}

\begin{figure*}
  \begin{center}
    \begin{tabular}{ll}
      \toprule
      \multicolumn{2}{c}{{\bf Correct top-1 guesses}} \\
      \cmidrule(r){1-2}
      {\bf Theorem statement} &
      \begin{lstlisting}
        ∀ {α : Type u_1} {β : Type u_2} [_inst_1 : decidable_eq α]
        [_inst_2 : decidable_eq β] (s : finset α) (t : finset β),
        s.product t = s.bUnion
        (λ (a : α), finset.image (λ (b : β), (a, b)) t)
      \end{lstlisting}
      \\
      {\bf Ground truth} &
                           \begin{lstlisting}
                             finset.product_eq_bUnion
                           \end{lstlisting}
      \\
      \cmidrule(r){1-2}
      {\bf Theorem statement} &
      \begin{lstlisting}
        ∀ {α : Type u_1} {β : Type u_2} [_inst_1 : topological_space α]
        [_inst_2 : topological_space β] {f : α → β},
        quotient_map f → function.surjective f
      \end{lstlisting}
      \\
      {\bf Ground truth} &
                           \begin{lstlisting}
                             quotient_map.surjective
                           \end{lstlisting}
      \\
            \cmidrule(r){1-2}
      {\bf Theorem statement} &
      \begin{lstlisting}
        ∀ {α : Type u_1} {β : Type u_2} (f : α → option β)
        (x : option α), x.pbind (λ (a : α) (_x : a ∈ x), f a) = x.bind f
      \end{lstlisting}
      \\
      {\bf Ground truth} &
                           \begin{lstlisting}
                             option.pbind_eq_bind
                           \end{lstlisting}
      \\
            \cmidrule(r){1-2}
      {\bf Theorem statement} &
      \begin{lstlisting}
        ∀ {C : Type u₁} [_inst_1 : category_theory.category C]
        {G : C ⥤ C} [_inst_2 : category_theory.comonad G]
        {A B : category_theory.comonad.coalgebra G} (h : A.A ≅ B.A)
        (w : A.a ≫ G.map h.hom = h.hom ≫ B.a),
        (category_theory.comonad.coalgebra.iso_mk h w).hom.f = h.hom
      \end{lstlisting}
      \\
      {\bf Ground truth} &
                           \begin{lstlisting}
                             category_theory.comonad.coalgebra.iso_mk_hom_f
                           \end{lstlisting}
      \\
            \cmidrule(r){1-2}
      {\bf Theorem statement} &
      \begin{lstlisting}
        ∀ {𝕜 : Type u_1} {E : Type u_2} [_inst_1 : is_R_or_C ,𝕜]
        [_inst_2 : inner_product_space 𝕜 E]
        [_inst_4 : normed_space ℝ E] [_inst_5 : is_scalar_tower ℝ 𝕜 E]
        (p x : E × E),
        ⇑(fderiv_inner_clm p) x =
          has_inner.inner p.fst x.snd + has_inner.inner x.fst p.snd
      \end{lstlisting}
      \\
      {\bf Ground truth} &
                           \begin{lstlisting}
                             fderiv_inner_clm_apply
                           \end{lstlisting}
      \\
      \bottomrule
    \end{tabular}
    \caption{A sample of correct top-1 guesses by our best model
      \texttt{wm-to-tt-m1-m2} on the {\em theorem naming}
      task. We performed this experiment on the
      \texttt{future-mathlib} evaluation set, which comprises
      entirely unseen theorems added to \texttt{mathlib} only after we
      last extracted training data.
  }
    \label{fig:naming-experiment-correct}
  \end{center}
\end{figure*}

\begin{figure*}
  \begin{center}
    \begin{tabular}{ll}
      \toprule
      \multicolumn{2}{c}{{\bf Incorrect guesses}} \\
      \cmidrule(r){1-2}
      {\bf Theorem statement} &
      \begin{lstlisting}
        ∀ {α : Type u_1} (t : ordnode α) (x : α),
        t.dual.find_min' x = ordnode.find_max' x t
      \end{lstlisting}
      \\
      {\bf Guesses (top 8)} &
\begin{lstlisting}
  ordinal.find_min'_eq, ordinal.find_min'_eq_max', ordinal.find_min'_def,
  ordinal.find_min'_eq_max, ordinal.find_min', ordinal.dual_find_min',
  ordinal.find_min'_gt, ordinal.find_min'_q
\end{lstlisting}
      \\
      {\bf Ground truth} &
                           \begin{lstlisting}
                             ordnode.find_min'_dual
                           \end{lstlisting}
      \\
      \cmidrule(r){1-2}
      {\bf Theorem statement} &
      \begin{lstlisting}
∀ {α : Type u_1} {β : Type u_3} {γ : Type u_5} [_inst_1 :
measurable_space α] [_inst_3 : measurable_space β]
[_inst_5 : measurable_space γ] {μ : measure_theory.measure α}
{ν : measure_theory.measure β}
[_inst_8 : measure_theory.sigma_finite ν]
{f : α × β → γ},
ae_measurable f (μ.prod ν) → (∀ᵐ(x : α) ∂μ,
  ae_measurable (λ (y : β), f (x, y)) ν)
      \end{lstlisting}
      \\
      {\bf Guesses (top 8)} &
\begin{lstlisting}
measure_theory.ae_prod, measure_theory.ae_of_ae_prod,
measure_theory.ae_eq_prod_of_ae, measure_theory.ae_ae_of_ae_prod,
measure_theory.ae_measure_prod_mk_left,
measure_theory.ae_prod_of_ae_prod,
measure_theory.ae_measure_prod, measure_theory.ae_eq_refl
\end{lstlisting}
      \\
      {\bf Ground truth} &
                           \begin{lstlisting}
                             ae_measurable.prod_mk_left
                           \end{lstlisting}
      \\

      \cmidrule(r){1-2}
      {\bf Theorem statement} &
      \begin{lstlisting}
        ∀ {α : Type u_1} {β : Type u_2} {γ : Type u_3}
        {f : filter α} {h : set α → set β} {m : γ → β}
        {l : filter γ}, filter.tendsto m l (f.lift' h) ↔
          ∀ (s : set α), s ∈ f → (∀ᶠ (a : γ) in l, m a ∈ h s)
      \end{lstlisting}
      \\
      {\bf Guesses (top 8)} &
\begin{lstlisting}
filter.tendsto_lift'_iff, filter.tendsto_lift'_def
\end{lstlisting}
      \\
      {\bf Ground truth} &
                           \begin{lstlisting}
                             filter.tendsto_lift'
                           \end{lstlisting}
      \\
      \cmidrule(r){1-2}
      {\bf Theorem statement} &
      \begin{lstlisting}
        ∀ {R : Type} [_inst_1 : comm_ring R]
        {d : ℤ} (f : ℤ√d →+* R),
        ↑(⇑(zsqrtd.lift.symm) f) = ⇑f zsqrtd.sqrtd
      \end{lstlisting}
      \\
      {\bf Guesses (top 8)} &
\begin{lstlisting}
zsqrtd.coe_lift_symm, zsqrtd.coe_lift.symm, zsqrtd.lift.coe_symm_apply,
zsqrtd.lift_symm_apply, zsqrtd.lift.coe_coe_symm, zsqrtd.lift.coe_symm_coe,
zsqrtd.lift.symm_coe_zsqrtd, zsqrtd.lift_symm_to_zsqrtd
\end{lstlisting}
      \\
      {\bf Ground truth} &
                           \begin{lstlisting}
                             zsqrtd.lift_symm_apply_coe
                           \end{lstlisting}
      \\
      \bottomrule
    \end{tabular}
    \caption{A sample of incorrect guesses by our best model
      \texttt{wm-to-tt-m1-m2} on the {\em theorem naming}
      task. We performed this experiment on the
      \texttt{future-mathlib} evaluation set, which comprises
      entirely unseen theorems added to \texttt{mathlib} only after we
      last extracted training data. Most of the top-8 guesses
      displayed in the above table are very similar to the ground
      truth, in some cases being equivalent up to permutation of
      underscore-separated tokens. Note that for the first example,
      the concept of \texttt{ordnode} was not in the training data
      whatsoever and all predictions are in the syntactically similar
      \texttt{ordinal} namespace.
  }
    \label{fig:naming-experiment-incorrect}
  \end{center}
\end{figure*}

We included {\em theorem naming} as part of the PACT task suite. By
\texttt{mathlib} convention, theorem names are essentially
snake-cased, natural language summaries of the type signature of
a theorem, and so the theorem naming task is analogous to a
formal-to-informal translation task. We evaluate the ability of our best
model (in terms of theorem proving success rate)
\texttt{wm-to-tt-m1-m2} on its ability to guess theorem names on the
completely unseen \texttt{future-mathlib} set of theorems. The
distribution shift inherent in the \texttt{future-mathlib} dataset
particularly impacts the theorem naming task, because many of the
ground-truth names will
involve names for concepts that were only defined in \texttt{mathlib}
{\em after} we
extracted our training data.

On the \(\approx\)\(2.8\)K
\texttt{future-mathlib} theorems, we queried \texttt{wm-to-tt-m1-m2}
for up to \(N = 16\) candidates. We order these candidates into a list
\texttt{xs} by
decreasing cumulative log-probability and calculate the top-\(K\)
accuracy by checking if any of the first \(K\) candidates of \texttt{xs} match the
ground truth exactly. The model \texttt{wm-to-tt-m1-m2} was able to
achieve 20.1\% top-1 accuracy, 21.1\% top-3 accuracy, 26.7\% top-10 accuracy, and 30.0\%
top-16 accuracy. We display a sample of correct top-1 guesses
(\Cref{fig:naming-experiment-correct}) and a sample of failed guesses in
(\Cref{fig:naming-experiment-incorrect}).
We note that the failed guesses, while containing no syntactic
matches, are both semantically reasonable and syntactically very
similar to the ground truth.

\subsection{Test set evaluation breakdown by module}

\begin{figure*}
  \begin{center}
     \includegraphics[width=0.9\textwidth]{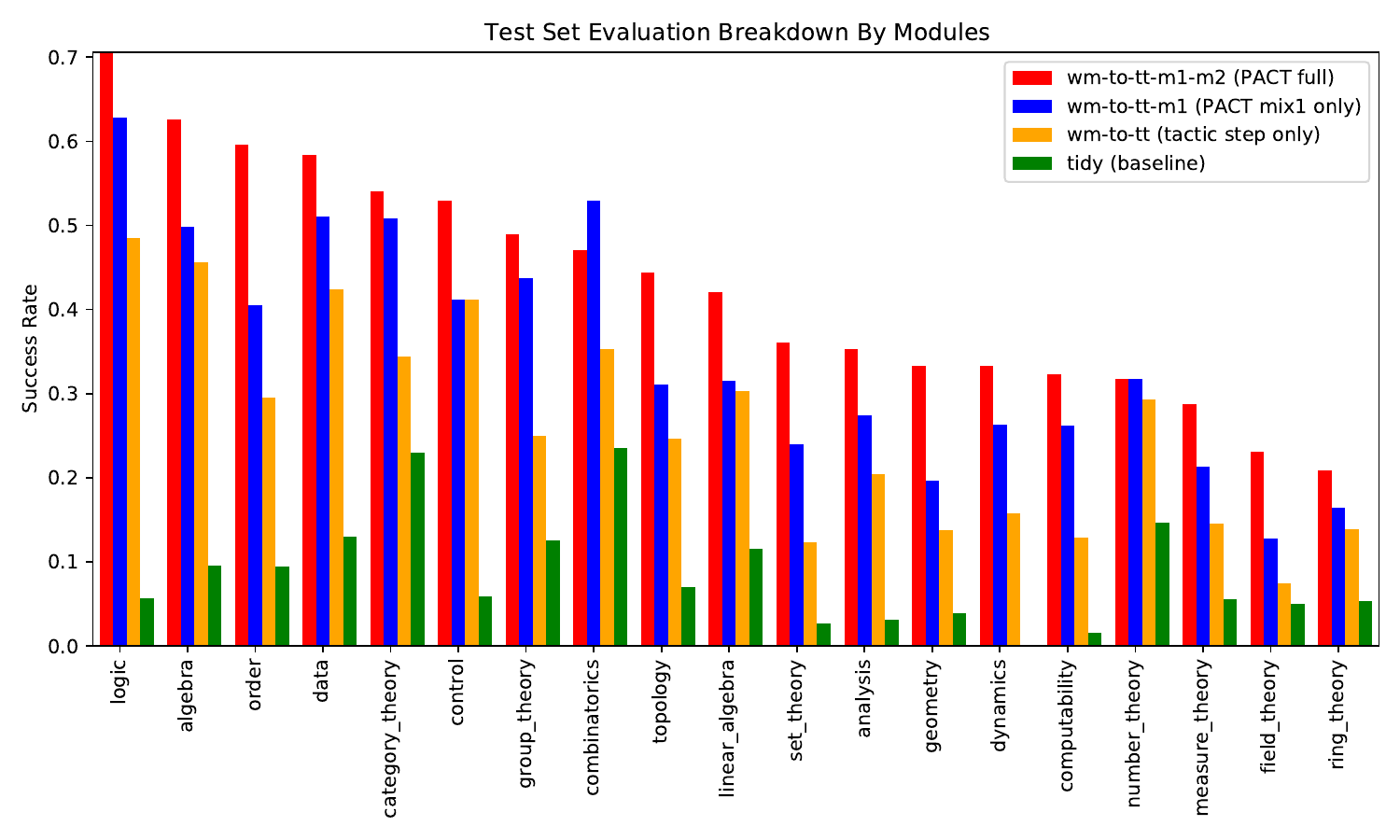}
     \caption{
       A breakdown of theorem proving success rate on the
       \texttt{test} set for
       \texttt{wm-to-tt-m1-m2}, \texttt{wm-to-tt-m1},
       \texttt{wm-to-tt}, and the \texttt{tidy} baseline across
       top-level modules in Lean's \texttt{mathlib}. We see that
       \texttt{wm-to-tt-m1-m2} mostly dominates \texttt{wm-to-tt-m1}
       and the models trained using PACT dominate the model
       \texttt{wm-to-tt} trained on human tactic proof steps.
}
    \label{fig:test-set-eval-bar-plot}
  \end{center}
\end{figure*}

Lean's \texttt{mathlib} is organized into top-level modules, which
roughly organize theorems into mathematical subject area. In
\Cref{fig:test-set-eval-bar-plot}, we break down the
evaluation results on our \texttt{test} set between our PACT-trained models
\texttt{wm-to-tt-m1-m2} and \texttt{wm-to-tt-m1} and our baselines
\texttt{wm-to-tt} and \texttt{tidy}. We see that full PACT mostly
dominates over co-training on just the \texttt{mix1} tasks over all
subject areas, and that \texttt{wm-to-tt-m1} dominates the model
\texttt{wm-to-tt} trained on human tactic proof steps only.

\subsection{Baseline description}\label{subsec:appendix_baseline}

The \texttt{tidy} backend is determined by a constant oracle
\begin{lstlisting}
  Ω : tactic_state → list (string × float)
\end{lstlisting}
which always returns the same list of tactics, namely:

\begin{lstlisting}
  meta def tidy_default_tactics : list (string × float) :=
  list.map (flip prod.mk 0.0) [
     "refl"
  ,  "exact dec_trivial"
  ,  "assumption"
  ,  "tactic.intros1"
  ,  "tactic.auto_cases"
  ,  "apply_auto_param"
  ,  "dsimp at *"
  ,  "simp at *"
  ,  "ext1"
  ,  "fsplit"
  ,  "injections_and_clear"
  ,  "solve_by_elim"
  ,  "norm_cast"
]
\end{lstlisting}

Unlike the \texttt{gptf} backend, which generates a list of candidates in
parallel independently, \texttt{tidy} enjoys the advantage that the
list of tactics it emits is carefully chosen and ordered in order to
optimize the proof search---this is based on the ``waterfall''
technique of the human-style
automated theorem prover described in~\citep{DBLP:journals/jar/GanesalingamG17}.

\subsection{Computational resource estimates}\label{subsec:appendix_compute}

For each evaluation loop over the \texttt{test} set, we
distributed the theorems over a pool of \(32\) CPU workers whose
inference requests were load-balanced over \(4\) \texttt{V100}
GPUs. Each evaluation required \(\approx\)\(10\) hours with \(\approx\)\(
30\%\) GPU utilization. We observed that our evaluation
was bottlenecked by inference and in practice, we hosted up to three
evaluation loops at once on a VM with 80 logical cores without
achieving full CPU utilization. In addition to the wall-clock timeout of 600s, we
also limited the proof search to a logical timeout of \texttt{512} iterations,
where one iteration corresponds to a single expansion of a node of the BFS
search tree. In practice, so much time was spent either blocked on
inference or performing the tactic executions in
the inner loop of each iteration that we rarely exceeded the logical timeout,
usually exceeding the wall-clock timeout instead.

Fine-tuning on our largest dataset \texttt{mix1 + mix2 + tactic} required
\(26\) hours using \(64\) \texttt{A100} GPUs exhibiting high \texttt{FP16}
usage, totalling an estimated \(\approx\)\(1.5\mathrm{K}\)
\texttt{A100(FP16)}-hours. This gives an estimated cost of \(17.33\)
\texttt{A100(FP16)}-hours per billion elapsed tokens during training.
We note that when calculating the number of elapsed tokens for
training, we overestimate the actual number of tokens effectively
trained on by summing full context windows (in this case, 2048 tokens).

\section{Example proofs}\label{sec:appendix_examples}
Lean's \texttt{mathlib} is one of the most active open-source software
projects in the world. More than one-third of the proofs found by our
models are shorter and produce smaller proof terms than the ground
truth, leading to dozens of GPT-f co-authored commits to
\texttt{mathlib}. We examine some of the proofs found by our models in
more detail.

\subsection{\texttt{lie\_algebra.morphism.map\_bot\_iff}}
This proof produces a proof term which is 4X smaller than the original:

\begin{lstlisting}
lemma map_bot_iff : I.map f = ⊥ ↔ I ≤ f.ker :=
by { rw ← le_bot_iff, apply lie_ideal.map_le_iff_le_comap }
\end{lstlisting}

The original, human-written proof is much longer, viz.
\begin{lstlisting}
lemma map_bot_iff : I.map f = ⊥ ↔ I ≤ f.ker :=
begin
  rw le_ker_iff, unfold lie_ideal.map, split; intros h,
  { rwa [eq_bot_iff, lie_submodule.lie_span_le, set.image_subset_iff, lie_submodule.bot_coe] at h,},
  { suffices : f '' I = ↑(⊥ : lie_ideal R L'), { rw [this, lie_submodule.lie_span_eq], },
    ext x, rw [lie_submodule.bot_coe, set.mem_singleton_iff, set.mem_image],
    split,
    { rintros ⟨y, hy, hx⟩, rw ← hx, exact h y hy, },
    { intros hx, use 0, simp [hx], }, },
end
\end{lstlisting}

\subsection{\texttt{primrec.of\_equiv}}

This proof produces a proof term which is 12X smaller than the original:

\begin{lstlisting}
theorem of_equiv {β} {e : β ≃ α} :
  by haveI := primcodable.of_equiv α e; exact
  primrec e :=
by letI : primcodable β := primcodable.of_equiv α e; exact encode_iff.1 primrec.encode
\end{lstlisting}

The author of the original proof and maintainer of that package commented:

\begin{quote}
\texttt{encode\_iff.1 primrec.encode} is clever, it's a way to translate \texttt{primrec} across an equivalence when the encode function is defined as \texttt{encode x = encode (e x)} where \texttt{e} is the isomorphism.
\end{quote}

As far as they knew, this trick was never used before in the \texttt{computability} package.

\subsection{\texttt{real.tan\_eq\_sin\_div\_cos}}

This proof
demonstrates our model's library knowledge and ability at
premise selection.

\begin{lstlisting}
lemma real.tan_eq_sin_div_cos (x : ℝ) : tan x = sin x / cos x :=
begin
  rw ← of_real_inj,
  simp only [complex.tan_eq_sin_div_cos, of_real_sin, of_real_cos, of_real_div, of_real_tan]
end
\end{lstlisting}

Our model was able to predict this entire list of \verb`simp` lemmas
in one shot. Note that the lemma \verb`complex.tan_eq_sin_div_cos`
in this list is the \emph{complex number} version of the result,
i.e. \texttt{{\ensuremath{\forall}} (x :\;{\ensuremath{\mathbb{C}}}),
  tan x = sin x / cos x}. The previous human-written version of the
proof did not use the more general version of the lemma on complex
numbers, demonstrating our model's ability to find more general cases
of lemmas. We contrast this with the human-written ground truth, which
is more complex and performs a case analysis using the complex cosine:

\begin{lstlisting}
  lemma tan_eq_sin_div_cos : tan x = sin x / cos x :=
if h : complex.cos x = 0 then by simp [sin, cos, tan, *, complex.tan, div_eq_mul_inv] at *
else
  by rw [sin, cos, tan, complex.tan, ← of_real_inj, div_eq_mul_inv, mul_re];
  simp [norm_sq, (div_div_eq_div_mul _ _ _).symm, div_self h]; refl
\end{lstlisting}

\subsection{\texttt{sym2.is\_diag\_iff\_proj\_eq}}

The proof of this lemma is longer than the ground truth and was not
contributed to \texttt{mathlib},
but we describe it here because the proof is original and includes a
nontrivial instantiation of an existential quantifier.

\begin{lstlisting}
theorem sym2.is_diag_iff_proj_eq (z : α × α) :
  is_diag ⟦z⟧ ↔ z.1 = z.2 :=
begin
    intros,
    simp only [is_diag, prod.ext_iff, quot.exists_rep, iff_true, not_true, eq_self_iff_true],
    simp [diag], split,
    { rintros ⟨y, hy⟩, cases hy; refl },
    intro h, cases z, existsi z_snd,
    cases h, refl,
end
\end{lstlisting}

Before \verb`existsi z_snd`, the goal state is

\begin{lstlisting}
z_fst z_snd: α
h: (z_fst, z_snd).fst = (z_fst, z_snd).snd
⊢ ∃ (y : α), (y, y) ≈ (z_fst, z_snd)
\end{lstlisting}

This goal state never appeared in \texttt{mathlib}.

\subsection{\texttt{norm\_le\_zero\_iff}}

The following proof is remarkable because it uses fewer tactic steps
and takes a different route to the proof than the
ground truth,
uses a complex idiom \lstinline{simpa [...] using @...},
and was predicted in one shot.

\begin{lstlisting}
lemma norm_le_zero_iff {α : Type u_1} [_inst_1 : normed_group α]
  {g : α} : ∥g∥ ≤ 0 ↔ g = 0 :=
by { simpa [le_antisymm_iff, norm_nonneg] using @norm_eq_zero α _ g }
-- ground truth:
-- by { rw[←dist_zero_right],
--      exact dist_le_zero }
\end{lstlisting}

The lemmas supplied between the square
brackets are used to simplify the main goal. The lemma supplied after
the keyword \lstinline{using} can further simplify the lemmas supplied
between the square brackets. The \lstinline{@} modifier makes all
arguments explicit. The string \lstinline{@norm_eq_zero} never
appeared in our training data but the prediction includes the correct
number of correctly typed arguments, and even replaces the second
argument with a placeholder \lstinline{_}, correctly guessing that it
can be inferred by the elaborator. Finally, this again showcases the
strength of our models as premise selectors: all three lemmas
\lstinline{le_antisymm_iff}, \lstinline{norm_nonneg}, and
\lstinline{norm_eq_zero} were not used in the human-supplied proof but
are necessary for this proof.

Moving forward, we hope that our neural theorem provers will continue to find ways
to improve \verb`mathlib` and assist in creating new proofs. More
generally, we hope neural theorem proving will one day be become a
routine part of the formalization workflow.


\end{document}